\documentclass[sigconf]{acmart}
\usepackage{verbatim}
\usepackage{graphicx}
\usepackage{amsmath}
\usepackage{balance}

\usepackage{booktabs} 
\usepackage{mathrsfs}
\usepackage{bm}
\usepackage{float}
\usepackage{multirow}
\usepackage{enumitem}
\usepackage[figuresright]{rotating}

\copyrightyear{2022} 
\acmYear{2022} 
\setcopyright{rightsretained} 
\acmConference[MM '22]{Proceedings of the 30th ACM International Conference on Multimedia}{October 10--14, 2022}{Lisboa, Portugal}
\acmBooktitle{Proceedings of the 30th ACM International Conference on Multimedia (MM '22), Oct. 10--14, 2022, Lisboa, Portugal}
\acmISBN{978-1-4503-9203-7/22/10}
\acmDOI{10.1145/3503161.3548023}

\AtBeginDocument{%
  }

\usepackage{etoolbox}
\makeatletter
\patchcmd{\maketitle}{\@copyrightpermission}{
   \begin{minipage}{0.3\columnwidth}
     \href{https://creativecommons.org/licenses/by/4.0/}{\includegraphics[width=0.90\textwidth]{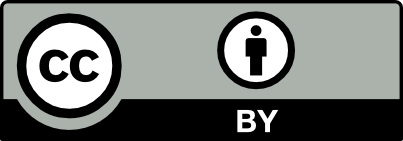}}
   \end{minipage}\hfill
   \begin{minipage}{0.7\columnwidth}
     \href{https://creativecommons.org/licenses/by/4.0/}{This work is licensed under a Creative Commons Attribution International 4.0 License.}
   \end{minipage}
  
   \vspace{5pt}
}{}{}

\makeatother

\begin{document}

\title[You Only Hypothesize Once]{You Only Hypothesize Once: \\ Point Cloud Registration with Rotation-equivariant Descriptors}

\author{Haiping Wang}
\authornote{Both authors contributed equally to this research.}
\orcid{0000-0002-8370-4585}
\affiliation{%
  \institution{Wuhan University}
  \streetaddress{No. 299, Bayi Road}
  \city{Wuhan}
  \state{Hubei}
  \country{China}
  \postcode{430072}
}
\email{hpwang@whu.edu.cn}

\author{Yuan Liu}
\authornotemark[1]
\orcid{0000-0003-2933-5667}
\affiliation{%
  \institution{The University of Hong Kong}
  \city{Hong Kong}
  \country{China}
}
\email{yliu@cs.hku.hk}

\author{Zhen Dong}
\authornote{The corresponding author.}
\orcid{0000-0002-0152-3300}
\affiliation{%
  \institution{Wuhan University}
  \city{Wuhan}
  \state{Hubei}
  \country{China}
}
\email{dongzhenwhu@whu.edu.cn}

\author{Wenping Wang}
\orcid{0000-0002-2284-3952}
\affiliation{%
  \institution{Texas A\&M University}
  \city{College Station}
  \state{Texas}
  \country{USA}
}
\email{wenping@tamu.edu}

\renewcommand{\shortauthors}{Haiping Wang, Yuan Liu, Zhen Dong, \& Wenping Wang}

\begin{abstract}
In this paper, we propose a novel local descriptor-based framework, called You Only Hypothesize Once (YOHO), for the registration of two unaligned point clouds. 
In contrast to most existing local descriptors which rely on a fragile local reference frame to gain rotation invariance, the proposed descriptor achieves the rotation invariance by recent technologies of group equivariant feature learning, which brings more robustness to point density and noise.
Meanwhile, the descriptor in YOHO also has a rotation-equivariant part, which enables us to estimate the registration from just one correspondence hypothesis. 
Such property reduces the searching space for feasible transformations, thus greatly improving both the accuracy and the efficiency of YOHO.
Extensive experiments show that YOHO achieves superior performances with much fewer needed RANSAC iterations on four widely-used datasets, the 3DMatch/3DLoMatch datasets, the ETH dataset and the WHU-TLS dataset.
More details are shown in our project page: https://hpwang-whu.github.io/YOHO/.
\end{abstract}





\begin{teaserfigure}
  \includegraphics[width=\textwidth]{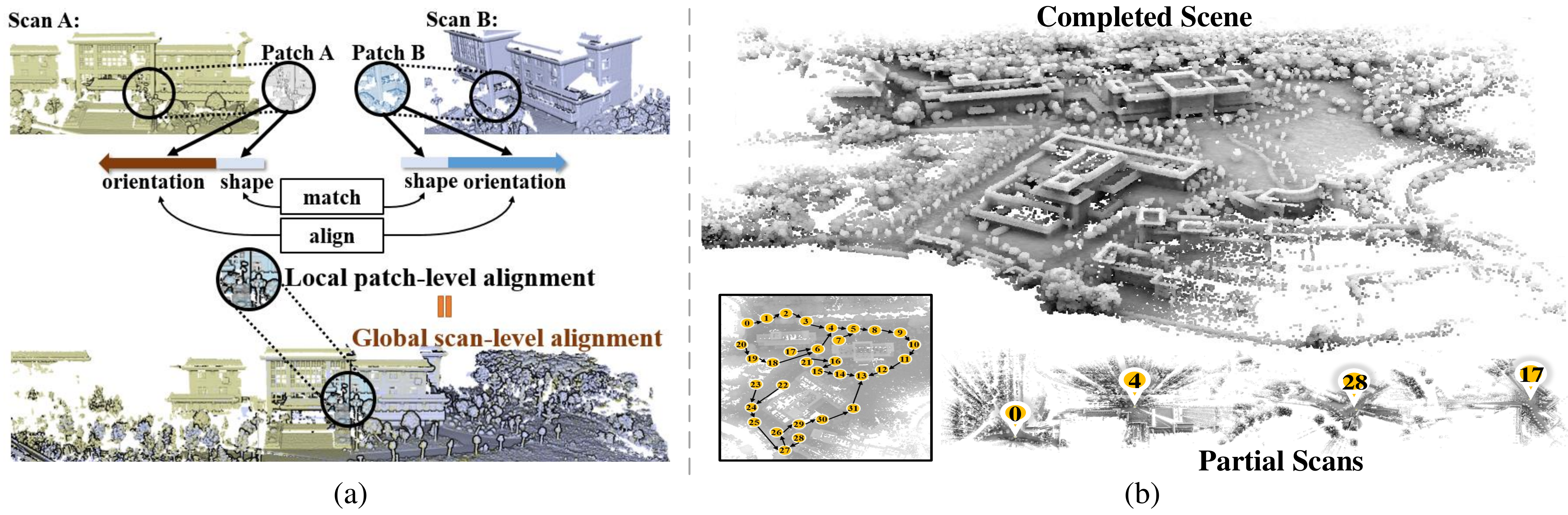}
  \caption{
    We present a novel framework called YOHO for point cloud registration. (a) The key idea of YOHO is to utilize orientations of local patches to find the global alignment of partial scans. (b) YOHO is able to automatically integrate partial scans into a completed scene, even these partial scans contain lots of noise and significant point density variations. 
  }
  \label{fig:teaser}
\end{teaserfigure}

\maketitle
\section{Introduction}
Finding an accurate SE(3) transformation of two unaligned partial point clouds, known as point cloud registration, is a prerequisite for many tasks such as 3D reconstruction~\cite{newcombe2011kinectfusion, gao2019sdm, zhong2019idfusion,liu2021votehmr}, pose estimation~\cite{aldoma2012tutorial,ge2015non,zhang2019danet,li2021lepard}, AR/VR~\cite{zhang2020deep, wang2018point, mahmood20193d,gao2019object}, and autonomous driving~\cite{yue2018lidar, mur2015orb, bailey2006simultaneous,wu2019ground}. 
Due to the large searching space of SE(3), directly finding an accurate transformation by enumerating is infeasible. A commonly-used pipeline usually consists of first extracting local descriptors of some interest points and then matching these local descriptors to build a set of putative correspondences. Hence, the searching space is only limited to SE(3) transformations that can explain the putative correspondences. However, due to noise and density variations in point clouds, the putative correspondences often contain lots of false correspondences so that the searching space still remains too large to find a correct transformation.

\begin{figure}
\begin{center}
\includegraphics[width=\linewidth]{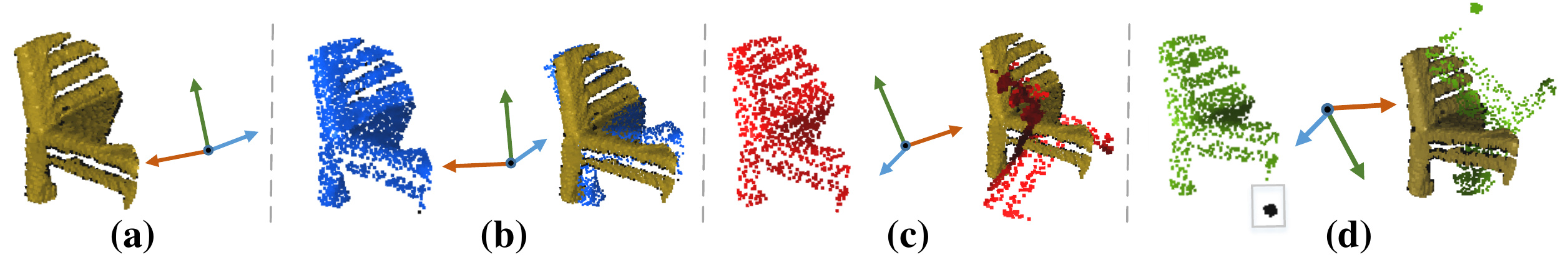}
\end{center}
\vspace{-10pt}
\caption{
LRF construction by PCA is sensitive to density variation or noise. (a) Original patch and its LRF. We Randomly downsample the point number to 2048 (b) and 1024 (c) and we add a small set of noise points (black) in (d), all of which lead to a obvious change of the LRF.
}
\label{fig:lrf}
\vspace{-10pt}
\end{figure}

In order to establish reliable correspondences, the local descriptor needs to be invariant to the rotation brought by the unknown SE(3) transformation. To achieve such invariance, most existing works~\cite{smooth,ao2020spinnet} align different interest points with local reference frames (LRF) constructed by principle component analysis (PCA)~\cite{wold1987principal}.
However, such a LRF construction is ambiguous~\cite{PCAamb} and usually sensitive to noise and point density as shown in Fig.~\ref{fig:lrf}, which may produce a completely incorrect alignment between two corresponding points. 
The incorrect alignment poses a great challenge for the subsequent feature extraction to find a rotation-invariant descriptor. 
Some other works~\cite{deng2018ppfnet,ppffold,pan2021robust} resort to handcrafted rotation-invariant features like point-pair angles or distances to construct a descriptor. 
However, these handcrafted features usually discard many useful geometry information thus are less discriminative. 
Recent technologies~\cite{cohen2016group,cohen2018spherical,EMVN,EPN,spherical,simeonov2021neural} of extracting features on the $SO(3)$ or $SE(3)$ group brings potentials for directly learning a rotation-invariant descriptor from raw point clouds, which takes good advantage of the powerful representation ability of neural networks to learn discriminative and rotation-invariant descriptors.

Though the nonalignment of two corresponding points poses a requirement of rotation invariance, it also provides an opportunity to estimate the rotation between two point clouds. This observation is illustrated in Fig.~\ref{fig:teaser} (a), where the rotation aligning the two corresponding 3D local patches is exactly the ground truth rotation for the registration of two point clouds. 
Based on this observation, a pair of matched descriptors, called rotation-equivariant descriptors, will have the ability to estimate a rotation if they encode such orientation information of the 3D local patches.
Only few existing registration works~\cite{deng20193d,dong2017novel} are based on the rotation-equivariant descriptors. These methods rely on the fragile LRF as the orientation~\cite{dong2017novel} or directly regress a rotation from an unaligned patch pair~\cite{deng20193d}, which does not generalizes well to unseen point pairs.

In this paper, we aim at designing a novel descriptor-based framework for point cloud registration, which simultaneously utilizes the rotation-invariance to find matches and the rotation-equivariance to find plausible rotations.
Instead of relying on a vulnerable external LRF, The rotation invariance is naturally achieved by feature extraction on the $SO(3)$ space via neural networks~\cite{cohen2016group,EMVN,EPN,DRgroup}, which takes advantage of powerful ability of neural networks to achieve robustness to noise and density variation of point clouds. 
Meanwhile, by utilizing the rotation equivariance, we can estimate the rotation from just one matched point pair. 
By combining the estimated rotation with the translation from the same matched point pair, we are able to use only one match pair as a hypothesis in RANSAC to estimate the global transformation.
In light of this, we call our framework \textit{You Only Hypothesize Once} (YOHO), which greatly improves both the accuracy and efficiency of registration by reducing the searching space for transformations.

Specifically, our descriptor, called YOHO-Desc, is built on a so-called group feature defined on the icosahedral group $G$, the largest finite subgroup of $SO(3)$, with $\bm{f}:G \to \mathbb{R}^n$. 
An essential observation on this group feature is that rotating the input point cloud with any rotations $g\in G$ will produce a new group feature $\bm{f}'$ which is only a permuted version of the original one: $\bm{f}'(h)=\bm{f}(gh)$, where $h\in G$. 
Three technical merits can be achieved. 
1) \textbf{Rotation estimation}. A relative rotation can be computed by finding a permutation in $G$ that aligns two group features.
2) \textbf{Rotation invariance}. The permutation can be simply eliminated by applying a pooling operator on the group feature, which results in a rotation-invariant descriptor. 
3) \textbf{Discriminative}. By applying a group-based convolution operator on the group feature $\bm{f}$, the descriptor is able to exploit discriminative patterns defined on the group $G$. 
To this end, modified RANSAC algorithms are designed to utilize both the estimated matches and the estimated rotations for accurate and efficient alignments.

We evaluate the performance of YOHO on the widely-used benchmarks 3DMatch~\cite{3dmatch} dataset, the 3DLoMatch~\cite{predator} dataset, the ETH \cite{smooth} dataset and the WHU-TLS~\cite{dong2020registration} dataset. The results show that YOHO achieves better or comparable accuracy of registration than the state-of-the-arts even with only 1000 RANSAC iterations, while existing methods commonly need 50000 iterations. The reduced RANSAC iterations greatly save time to reconstruct a completed point cloud from partial scans.

Our contributions are summarized as follows:
\begin{enumerate}[topsep=0pt,itemsep=-0.8ex,partopsep=0.8ex,parsep=1ex,leftmargin=2em]
\item We propose a novel rotation-equivariant local descriptor for point cloud registration, which is built on the icosahedral group features and general to combine with fully convolutional or patch-based backbones for promising enhancements.
\item We utilize the rotation-equivariance of the descriptor for the rotation estimation which greatly reduces the searching space for true transformations.
\item We design a registration framework YOHO, which achieves superior performance on benchmarks with less than 1k RANSAC iterations.
\end{enumerate}

\section{Related work}

\subsection{Point cloud registration}

\textbf{Feature-based methods}. Using local descriptors for point cloud registration has a long history\cite{diez2015qualitative,dong2020registration,learning3d}. Traditional methods~\cite{fourpcs,rusu2009fast,salti2014shot,dong2017novel,guo2013rops,jung2020geometric} use handcrafted features to construct descriptors. After the prosperity of deep learning models, learning based descriptors~\cite{smooth,FCGF,d3feat,huang20193d,ppffold,li2020end,lu2020rskdd} achieve impressive improvements over traditional handcrafted descriptors. Most of these methods achieve rotation invariance by PCA on the neighborhood to find one axis~\cite{mitra2003estimating,ao2020spinnet}, i.e. normal, or three axes directly~\cite{smooth,sun2019srinet,d3feat}. Some other works~\cite{ppffold,deng2018ppfnet} handcraft some rotation-invariant features and then use a network to extract descriptors from these  invariant features. However, the PCA on the neighborhood is not robust to noise or density variations and handcrafted invariant features usually lose lots of information.

Recent works~\cite{spherical,EPN} resort to group convolution layers~\cite{cohen2016group} and pooling on the group to extract rotation-invariant descriptors for point cloud registration. The most similar one is EPN~\cite{EPN}, which also adopts icosahedral group convolution for point cloud registration. The key difference is that YOHO proposes to utilize the estimated rotation in the modified RANSAC for fast and accurate partial scan alignment while EPN does not consider such rotations in the RANSAC, which leads to less efficient and less accurate results as demonstrated in experiments. We provide more discussion about EPN and YOHO in appendicess.

There are only few existing works~\cite{dong2017novel,deng20193d,zhu2022correspondence} estimate a rotation from a single descriptor pair for point cloud registration. The most relevant work is RelativeNet~\cite{deng20193d} which direct regress a rotation from a descriptor pair. However, such regression does not generalize well to unseen data in training set as shown by experiments. In contrast, by utilizing feature map defined on the icosahedral group, YOHO is able to estimate a plausible rotation by finding a permutation to align two feature maps, which improves the generalization ability of YOHO-Desc to unseen data. 
Meanwhile, YOHO-Desc achieves both rotation invariance and equivariance in a compact framework while RelativeNet relies on a separated PPF-FoldNet~\cite{ppffold} to construct rotation-invariant descriptors.

\textbf{Direct registration methods}. Instead of feature extraction on two point clouds separately, some other works find the alignment by simultaneously considering information from both two point clouds~\cite{icp,SIGicp,sparseicp,yang2015go}. These works either directly solve for the transformation parameters by networks~\cite{huang2020feature,aoki2019pointnetlk,li2020deterministic,yuan2020deepgmr,zhou2016fast} or estimate more accurate correspondences by conditioning descriptors of one point cloud on the other point cloud~\cite{predator,lu2019deepvcp,qiao2020end,pais20203dregnet,choy2020deep,li2019iterative,wang2019prnet,wang2019deep,fu2021robust,bai2021pointdsc,cofi}. In general, more accurate correspondences can be found due to additional information from the other point cloud. However, YOHO does not belong to this category because YOHO-Desc is constructed separately in two point clouds and the correspondences are established simply by the nearest neighbor matching. 

\subsection{Equivariant feature learning}
Recent works~\cite{cohen2016group,cohen2016steerable,cohen2018spherical,esteves2018learning,esteves2020spin,carlos2018polar,worrall2017harmonic,VN,poulenard2021functional} develop tools to learn equivariant features. Some works~\cite{DRgroup,zhao2020quaternion,sun2020canonical,poulenard2019effective,yu2020deep,kim2020rotation,shen20193d,EMVN,EPN} design architectures to learn a rotation-equivariant or -invariant features on point clouds. Recent Neural Descriptor Field~\cite{simeonov2021neural} applies equivariance~\cite{VN} to learn category-level descriptors with self-supervision. YOHO focuses on applying these rotation-equivariant feature learning techniques for the general point cloud registration task. The most similar works are \cite{DRgroup,EMVN,EPN}. These works mainly focus on the rotation-invariance for point cloud recognition or shape description. In comparison, YOHO simultaneously takes advantage of both the rotation equivariance and the rotation invariance for robust and effective point cloud registration.

\begin{figure}
\begin{center}
\includegraphics[width=\linewidth]{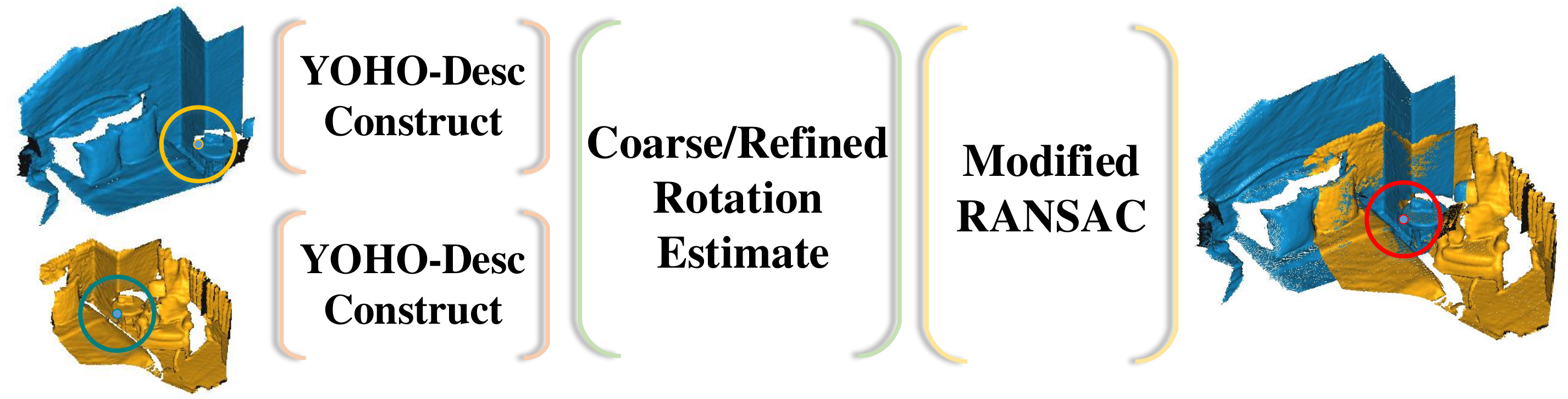}
\end{center}
   \caption{Overview of YOHO. YOHO-Desc is constructed on point cloud separately and is matched to build correspondences. On every correspondence, a coarse rotation and a refined rotation are estimated, which are further utilized by modified RANSAC algorithms to find correct transformations.}
\label{fig:overview}
\vspace{-10pt}
\end{figure}

\section{Method}
\textbf{Overview}. Given two unaligned point clouds $\mathcal{P}$ and $\mathcal{Q}$, our target is to find a transformation $\bm{T}=\{\bm{R},\bm{t}\}\in SE(3)$ that can integrate them into a whole point cloud. The whole pipeline of YOHO is shown as Fig.~\ref{fig:overview}. In the following, we first introduce the background in Sec.~\ref{sec:back}. Then, we introduce how to extract YOHO-Desc on a point cloud in Sec.~\ref{sec:desc}. The extracted YOHO-Desc will be matched to produce correspondences and we introduce how to estimate a rotation on a correspondence via YOHO-Desc in Sec.~\ref{sec:rot}. Finally, we utilize the estimated rotations in two modified RANSAC algorithms to find the transformations in Sec.~\ref{sec:ransac}.

\begin{figure}
\begin{center}
\includegraphics[width=\linewidth]{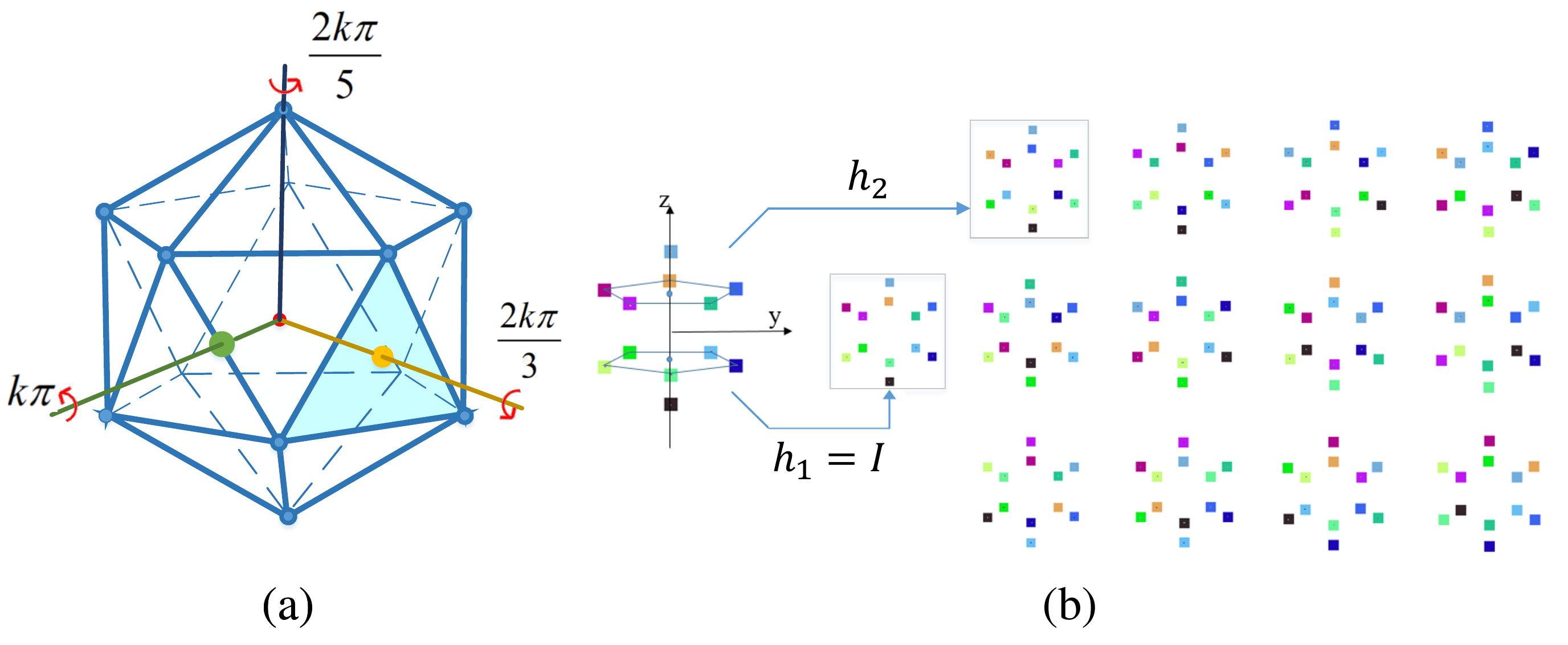}
\end{center}
	\caption{(a) Icosahedral group contains k$\pi$ rotations around axes through edge centers or 2k$\pi$/3 rotations around axes through face centers or 2k$\pi$/5 rotations around axes through vertices. (b) The neighborhood set $H$. Different colors are drawn on the vertices of an icosahedron. $H$ contains the identity element and 12 rotations that permute the identity vertices to the right 12 vertices. Note all 12 rotations are 72$^\circ$ rotations about axes.}
\label{fig:Igroupnei}
\vspace{-10pt}
\end{figure}
\text

\vspace{-10pt}
\subsection{Preliminary}
\label{sec:back}

In this section, we only introduce some backgrounds about $SO(3)$ space and recommend readers refer to \cite{cohen2016group,EMVN,liu2019gift}.

\textbf{Icosahedral group}. We define feature maps on the largest discrete finite subgroup of $SO(3)$, i.e. icosahedral group $G$. The icosahedral group consists of 60 rotations that keep a regular icosahedron invariant, as shown in Fig.~\ref{fig:Igroupnei}. Due to the closure of a group, $\forall g\in G$ and $\forall h \in G$, we have $gh\in G$, where $gh$ means the composition of the rotation $g$ and the rotation $h$ to get a new rotation. 

\textbf{Group action}. We can use an element $g\in G$ to act on other objects. Two kinds of group actions are used: $T_g \circ \mathcal{P}$ means rotating a set of points $\mathcal{P}$ by the rotation $g\in G$ and $P_g \circ \bm{f}$ means permuting the matrix $\bm{f}$ according to $g\in G$, which we will give a more detailed description in Sec.~\ref{sec:desc}. 

\textbf{Neighborhood set}. In order to define the convolution layer on the icosahedral group $G$, we define a neighborhood set $H$ as shown in Fig.~\ref{fig:Igroupnei}, which is similar to the $3\times 3$ or $5\times 5$ neighborhood in the vanilla image convolution.

\begin{figure}
\begin{center}
\includegraphics[width=1\linewidth]{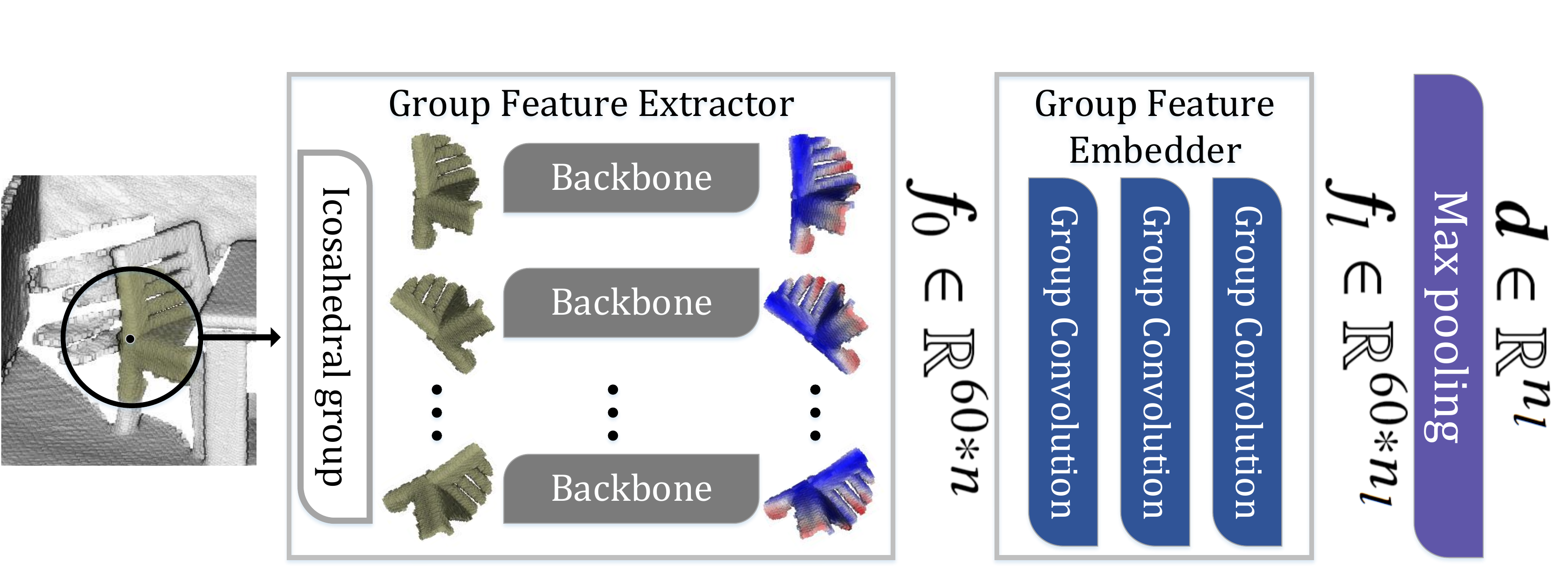}
\end{center}
\vspace{-3pt}
\caption{Pipeline for the descriptor construction.}
\label{fig:desc}
\vspace{-10pt}
\end{figure}

\subsection{Descriptor construction}
\label{sec:desc}
For a point $\bm{p} \in \mathcal{P}$, we construct a YOHO-Desc from its local 3D patch $N_{\bm{p}}=\{\bm{p}_i| \|\bm{p}_i-\bm{p}\|< r \}$, by two modules called Group Feature Extractor and Group Feature Embedder, as shown in Fig.~\ref{fig:desc}.

\textbf{Group feature extractor}. Given the input neighborhood point set $N_{\bm{p}}$, we rotate it with an element $g$ in the icosahedral group $G$. Every rotated point set is processed by the same point set feature extractor called \textit{backbone} to extract a $n$-dimensional feature, which can be expressed by
\begin{equation}
    \bm{f}_0(g)=\phi(T_g \circ N_{\bm{p}}),
\end{equation}
where $\bm{f}_{0}:G \to \mathbb{R}^{n_0}$ is the output group feature for the point $\bm{p}$, $\phi$ is the backbone and $T_g\circ N_{\bm{p}}$ means rotating the point set $N_{\bm{p}}$ with the rotation $g$. Since the icosahedral group $G$ has 60 rotations, the output group feature $\bm{f}_0$ is actually stored by a $60 \times n_0$ matrix where the row index stands for different rotations in $G$. Note any point set feature extractor, including PointNet~\cite{PN} and fully convolutional extractor FCGF~\cite{FCGF} or D3Feats~\cite{d3feat}, can be used as the backbone. When using the lightweight fully convolution based extractor as the backbone, we directly rotate the whole point cloud and the neighborhood set is implicitly defined by convolution operators.

\textbf{Group feature embedder}. The group feature can be processed by a localized icosahedral group convolution~\cite{EMVN} as follows,
\begin{equation}
    \label{eq:gconv}
    [\bm{f}_{k+1}(g)]_j=\sum_i^{13} \bm{w}_{j,i}^T \bm{f}_{k}(h_i g)+b_j,
\end{equation}
where $k$ is the index of the layers, $\bm{f}_{k}(g)\in \mathbb{R}^{n_k}$ and $\bm{f}_{k+1}(g)\in \mathbb{R}^{n_{k+1}}$ are the input and output feature respectively, $[\cdot]_j$ means the $j$-th element from the vector, $h_i\in H$ are all elements from the neighborhood set $H$, $\bm{w}_{j,i} \in \mathbb{R}^{n_k}$ is a trainable $n_k$-dimensional weight defined on the $h_i$ and $b_j$ is a trainable bias term. Note that the $j=1,...,n_{k+1}$ is the index for the output feature dimension and the composition $h_ig$ is also an element in $G$ due to the closure property. Such a group convolution layer enables us to exploit the local patterns defined on the icosahedral group. To this end, $l$ group convolution layers defined in Eq.~\ref{eq:gconv} along with subsequent ReLU and Batch Normalization layers are stacked to construct the group feature embedder.

\textbf{Rotation equivariance}. An important property on the group feature $\bm{f}_k$ is that using a rotation $h\in G$ to rotate the input point set $N_{\bm p}$ will only result in a permuted version $\bm{f}'_k$ of the original group feature $\bm{f}_k$, which is
\begin{equation}
    \label{eq:equi}
    \bm{f}'_k=P_h \circ \bm{f}_k,
\end{equation}
where both $\bm{f}'_k$ and $\bm{f}_k$ are group features represented by $60 \times n_k$ matrices and $P_h \circ \bm{f}_k$ is actually a permutation of row vectors of the matrix $\bm{f}_k$ brought by $h$. Note Eq.~\ref{eq:equi} holds for all group features $k=0,1,...,l$. We leave the proof of Eq.~\ref{eq:equi} in the appendices.

\textbf{Invariant descriptor}. Based on the equivariance property, we can construct a rotation-invariant descriptor from the final layer group feature $\bm{f}_l$ by simply applying an average-pooling operator on all group elements, which is,
\begin{equation}
    \bm{d}={\rm AvgPool}(\bm{f}_l). 
\end{equation}
The resulted descriptor $\bm{d}$ is invariant to all rotations in the icosahedral group, which can be easily verified that $\bm{d}'={\rm AvgPool}(P_h \circ \bm{f}_l)={\rm AvgPool}(\bm{f}_l)=\bm{d}$ since $\rm AvgPool$ is unaffected by permutations. Though $\bm{d}$ is not strictly invariant to rotations outside the icosahedral group, the existing rotation invariance already provides a strong inductive bias for the network to learn invariance to other rotations.

\subsection{Rotation estimation}
\label{sec:rot}

We apply a nearest neighborhood (NN) matcher on the descriptors $\bm{d}$ with a mutual nearest check~\cite{3dmatch,predator,ao2020spinnet} from two scans to find a set of putative correspondences. In this section, we introduce how to compute a rotation on every correspondence.

\textbf{Coarse rotations}. Given two matched points $\bm p$ and $\bm q$ along with their YOHO-Desc, we find a coarse rotation $R_c$ by aligning two group features with a permutation
\begin{equation}
    R_c=\mathop{argmin}_{g \in G} \|\bm{f}_{l,\bm{q}}-P_{g} \circ \bm{f}_{l,\bm{p}}\|_2,
    \label{eq:coarse}
\end{equation}
where $\bm{f}_{l,\bm{p}}$ is the final group feature for the point $\bm{p}$ and we iterate all 60 permutations in $\{P_g | g\in G\}$ to find the best one that minimizes the L2 distance between group features.

\begin{figure}
\begin{center}
\includegraphics[width=0.92\linewidth]{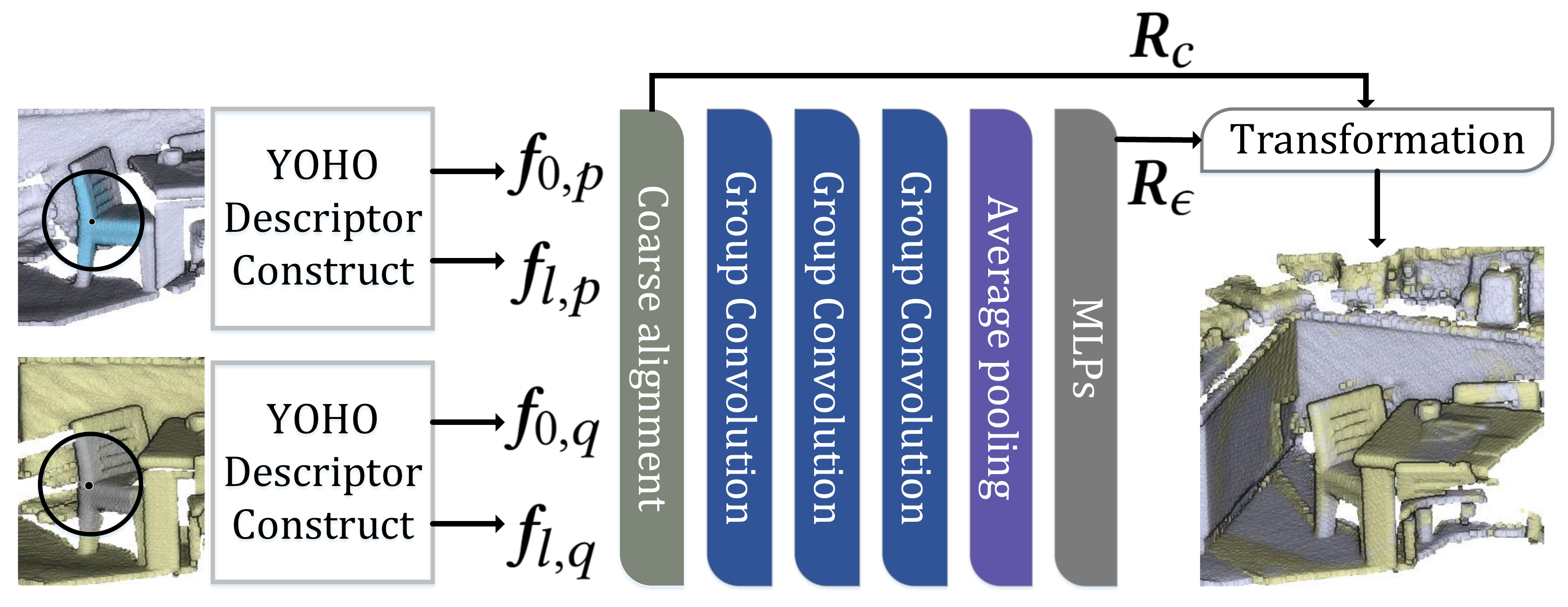}
\end{center}
\caption{Pipeline for the rotation estimation.}
\label{fig:rot}
\vspace{-10pt}
\end{figure}

\textbf{Refined rotations}. Since the group $G$ is a discretization of $SO(3)$, we cannot directly get a precise rotation by finding the best permutation. To find such a precise rotation, we use a regressor to compute the residual between the coarse rotation and the true rotation by
\begin{equation}
    R_\epsilon=\eta ([\bm{f}_{0,\bm{q}};\bm{f}_{l,\bm{q}};P_{R_c} \circ \bm{f}_{0,\bm{p}}; P_{R_c} \circ \bm{f}_{l,\bm{p}}]),
\end{equation}
where $R_\epsilon$ is the rotation residual, $[\cdot;\cdot]$ means the concatenation of features along the channel direction, $P_{R_c}\circ \bm{f}$ means the permutation of $\bm{f}$ by $R_c$, the resulted feature map by the concatenation is still a group feature with size $60\times(2n_0+2n_l)$ and the $\eta$ is a network. Note both the final group feature $\bm{f}_l$ and the initial group feature $\bm{f}_0$ are fed to the network $\eta$. $\eta$ is called Rotation Residual Regressor, which applies group convolutions defined by Eq.~\ref{eq:gconv} on the input group feature. After being processed by the group convolutions, the group feature is average-pooled to a single feature vector which is further processed by a Multi-Layer Perceptron (MLP) to regress a rotation residual in the quaternion form. The whole network is shown in Fig.~\ref{fig:rot}. The rotation residual $R_\epsilon$ is composited with the coarse rotation $R_c$ to produce the refined rotation $R_r=R_\epsilon R_c$.


\subsection{Modified RANSAC algorithms}
\label{sec:ransac}
YOHO descriptors are extracted separately on two point clouds (Sec.~\ref{sec:desc}) and are matched by a NN matcher to produce a set of correspondences $C=\{c_i=(\bm{p}_i,\bm{q}_i)|\bm{p}_i \in \mathcal{P}, \bm{q}_i \in \mathcal{Q}\}$. On every correspondence $c_i$, a coarse rotation $R_{c,i}$ and a refined rotation $R_{r,i}$ are computed as stated in Sec.~\ref{sec:rot}.
In a standard RANSAC algorithm, a correspondence triplet is randomly selected to compute a transformation $\bm{T}$ and the transformation with the largest number of inliers will be selected as the output transformation.
In our case, we propose two different ways, called Coarse Rotation Verification (CRV) and One-Shot Transformation Estimation (OSE), to incorporate the estimated rotations in the RANSAC pipeline, which greatly reduces the searching space for transformations.

\textbf{Coarse rotation verification}. Instead of randomly selecting all correspondence triplets to compute $\bm{T}$, we limit feasible correspondence triplets to $\{(i,j,k)|R_{c,i}=R_{c,j}=R_{c,k}\}$ that have the same estimated coarse rotations. Such a verification greatly reduces the searching space to find an accurate transformation.

\textbf{One-shot transformation estimation}. Given a correspondence $c_i$ and its refined rotation $R_{r,i}$, we can compute the transformation $\bm{T}$ directly by $\bm{R}=R_{r,i}$ and $\bm{t}=\bm{q}_i-\bm{R}\bm{p}_i$. Given $n$ correspondences, we will only have $n$ feasible transformation hypotheses which are much less than the $n\times(n-1)\times(n-2)/6$ feasible triplets in the original RANSAC. Moreover, if the inlier ratio in the putative correspondences is $\alpha$, then we will have a chance of $\alpha$ to find the true transformation while in the original RANSAC, such chance is $\alpha^3$.

\subsection{Implementation details}
By default, we implement YOHO using FCGF~\cite{FCGF} as the backbone. In the appendices, we also provide results with PointNet~\cite{PN} as the backbone to show YOHO's ability to work with different backbones. The group convolution network in the group feature embedding has 4 group convolution layers. The final descriptor $\bm{d}$ and the group feature $\bm{f}_l$ used before coarse rotation estimation are all normalized by their L2 norms. 
The rotation residual regressor has 3 group convolution layers and 3 layers in the MLP after the average-pooling layer.
A more detailed implementation about the network architecture can be found in our project page. 
To build the putative correspondences, YOHO-Desc are matched by a nearest neighborhood matcher with a mutual nearest test, which is exactly the same as used in~\cite{smooth}.

\textbf{Loss for descriptor construction}. 
Given a batch of ground-truth point pairs $\{(\bm{p},\bm{p^+})\}$ as well as their ground-truth rotations $\{R_{\bm{p}}\}$, we compute the outputs of group feature embedder, which are the rotation invariant descriptors $\{(\bm{d}_{\bm{p}},\bm{d}^+_{\bm{p}})\}$, the rotation equivariant group features $\{(\bm{f}_{\bm{p}},\bm{f}^+_{\bm{p}})\}$, and the corresponding ground truth coarse rotations $\{g^+_{\bm{p}}\}$.
For every sample in the batch, we compute the loss:
\begin{equation}
    \ell_1(\bm{d},\bm{d}^+,\bm{D}^-)=\frac{e^{||\bm{d}-\bm{d}^+||_2}-\underset{\bm{d}^- \in \bm{D}^-}{\min}e^{||\bm{d}-\bm{d}^-||_2}}{e^{||\bm{d}-\bm{d}^+||_2}+\underset{\bm{d}^- \in \bm{D}^-}{\sum}e^{||\bm{d}-\bm{d}^-||_2}}
    \label{invloss}
\end{equation}
\begin{equation}
    \ell_2(\bm{f},\bm{f}^+,g^+)=-log(\frac{e^{\left<\bm{f},P_{g^+}\circ \bm{f}^+\right>}}{\underset{g \in G}{\sum}e^{\left<\bm{f},P_{g}\circ \bm{f}^+\right>}})
    \label{eqvloss}
\end{equation}
\begin{equation}
    \ell_d=\lambda*\ell_1(\bm{d},\bm{d}^+,\bm{D}^-)+\ell_2(\bm{f},\bm{f}^+,g^+),
\end{equation}
where the subscript $\bm{p}$ is omitted for simplicity. Eq.~\ref{invloss} is used for the supervision of the rotation invariant descriptor, $\bm{d}$ is a rotation invariant descriptor, $\bm{d}^+$ is its matched descriptor, $\bm{D}^-$ are the other negative descriptors in the batch and $||.||_2$ is the L2 norm. 
Eq.~\ref{eqvloss} is used for the supervision of the coarse rotation estimation. $\bm{f}$ is a flattened query group feature vector, $\left<.,.\right>$ is the vector dot product and $g^+$ is the ground truth coarse rotation between $\bm{f}$ and $\bm{f}^+$.
$\lambda$ is set to 5. $\ell_1$ is the batch-hard loss while $\ell_2$ encourage the alignments of two group features under the ground-truth rotations. The reason to use $\ell_2$ is that though the equivariance property is theoretically guaranteed, some noise or density variations may break the equivariance so that supervision from $\ell_2$ will make the network more robust to these factors.

\textbf{Loss for Rotation Residual Regresser}.
Given a ground-truth point pair $(\bm{p},\bm{p^+})$ in the batch and its ground-truth rotation $R_{\bm{p}}$. We extract the group features using the group feature extractor and embedder. Then, we coarsely align the group features using the ground truth coarse rotation $g^+_{\bm{p}}$. Finally, we estimate its residual rotation $R_{\epsilon,\bm{p}}$ using the Rotation Residual Regresser and supervise the Regresser by:
\begin{equation}
    \ell_R(\bm{p},\bm{p^+})=\|R_{\epsilon,\bm{p}}-R_{\epsilon,\bm{p}}^+\|_2
    \label{eq:Rloss}
\end{equation}
where $R_{\epsilon,\bm{p}}^+=R_{\bm{p}}g_{\bm{p}}^{+T}$ is the ground truth residual rotation. $R_{\epsilon,\bm{p}},R_{\epsilon,\bm{p}}^+$ are represented in the quaternion form.

\section{Experiments}
In the following, we evaluate two YOHO models for point cloud registration, which are the YOHO using coarse rotation verification (YOHO-C) and the YOHO with the one-shot transformation estimation (YOHO-O).

\subsection{Experimental protocol}

\textbf{Datasets}. We follow exactly the same experiment protocol as \cite{3dmatch} to prepare the training and testing data on the indoor 3DMatch dataset. In this setting, 5000 predefined keypoints are extracted on every scan. 
However, since the original testset on 3DMatch only contains scan pairs with $>$30\% overlap, we also evaluate the model on the 3DLoMatch dataset~\cite{predator} with overlap between 10\% and 30\% to demonstrate our robustness to low overlap. 
We also evaluate the proposed method on the outdoor ETH dataset~\cite{smooth} and the outdoor WHU dataset~\cite{dong2020registration}, which contain more point density variations than the 3DMatch. Note that our model is trained on the 3DMatch dataset and is solely evaluated on the ETH dataset, the WHU-TLS dataset and the 3DLoMatch dataset. 
For the FCGF backbone, the downsampled voxel size is 2.5cm, 15cm and 0.8m for the 3DMatch, ETH and WHU-TLS datasets according to their scales.

\textbf{Metrics}. Feature Matching Recall (FMR)~\cite{smooth,FCGF,predator}, Inlier Ratio (IR)~\cite{predator} and Registration Recall (RR)~\cite{FCGF,d3feat,predator} are used as metrics for evaluations. A correspondence is regarded as a correct one if the distance between its two matched points is $\le \tau_{c}$ under the ground truth transformation. FMR is the percentage of scan pairs with correct correspondence proportions more than 5\% found by the local descriptor, same as used in \cite{ao2020spinnet,predator}. IR is the average correct correspondence proportions. RR is the percentage of correctly aligned scan pairs, which means that the average distance between the points under the estimated transformation and these points under the ground truth transformation is less than $\tau_{r}$. 
For RR, we follow \cite{predator} and report the number of RANSAC iterations to achieve the performance. Due to the randomness of RANSAC, we run YOHO three times to compute the mean of RR on all datasets.

\textbf{Baselines}. We mainly compare YOHO with state-of-the-art learning based descriptors: PerfectMatch~\cite{smooth}, FCGF~\cite{FCGF},  D3Feat~\cite{d3feat}, LMVD \cite{li2020end}, EPN~\cite{EPN} SpinNet~\cite{ao2020spinnet}, and a learning-based matcher: Predator~\cite{predator}. On the 3DMatch dataset, we also include the results from RelativeNet~\cite{deng20193d} which also learns rotation equivariance for registration.  For all baseline methods, we report the results in their papers or evaluate with their official codes or models.
For a fair comparison, baseline methods uses the RANSAC implementation in Open3D~\cite{zhou2018open3d} with engineering optimization like concurrent computation, optimized hyperparameters and distance checks.

\begin{table}[]
\begin{center}
\resizebox{\linewidth}{!}{
\begin{tabular}{lcccccc}
\toprule[1.3pt]
\multirow{2}{*}{}       & \multicolumn{1}{l}{} & \multicolumn{2}{c}{3DMatch}   & \multicolumn{2}{c}{3DLoMatch} \\
                        & \multicolumn{1}{l}{} & Origin        & Rotated       & Origin        & Rotated       \\ \hline
                        & \multicolumn{1}{l}{} & \multicolumn{4}{c}{Feature Matching Recall (\%)}              \\ \hline
RelativeNet\cite{deng20193d}& \multicolumn{1}{l}{} & 74.6          & -             & -             & -             \\
PerfectMatch\cite{smooth}    & \multicolumn{1}{l}{} & 95.0          & 94.9          & 63.6          & 63.4          \\
FCGF\cite{FCGF}              & \multicolumn{1}{l}{} & 97.4          & 97.6          & 76.6          & 75.4          \\
D3Feat\cite{d3feat}          & \multicolumn{1}{l}{} & 95.6          & 95.5          & 67.3          & 67.6          \\
LMVD\cite{li2020end}         & \multicolumn{1}{l}{} & 97.5          & 96.9          & \underline{78.7}    & \underline{78.4}    \\
EPN\cite{EPN}                & \multicolumn{1}{l}{} & \underline{97.6}    & \underline{97.6}    & 76.3          & 76.6            \\
SpinNet\cite{ao2020spinnet}  & \multicolumn{1}{l}{} & \underline{97.6}    & 97.5          & 75.3          & 75.3          \\
Predator\cite{predator}      & \multicolumn{1}{l}{} & 96.6          & 96.7          & 78.6          & 75.7          \\
YOHO                         & \multicolumn{1}{l}{} & \textbf{98.2} & \textbf{98.1} & \textbf{79.4} & \textbf{79.2} \\ \hline

                        & \multicolumn{1}{l}{} & \multicolumn{4}{c}{Inlier Ratio (\%)}              \\ \hline
PerfectMatch\cite{smooth}    & \multicolumn{1}{l}{} & 36.0          & 35.8          & 11.4          & 11.7          \\
FCGF\cite{FCGF}              & \multicolumn{1}{l}{} & 56.8          & 56.2          & 21.4          & 21.6          \\
D3Feat\cite{d3feat}          & \multicolumn{1}{l}{} & 39.0          & 39.2          & 13.2          & 13.5          \\
LMVD\cite{li2020end}         & \multicolumn{1}{l}{} & 45.1          & 45.0          & 17.3          & 17.0          \\
EPN\cite{EPN}                & \multicolumn{1}{l}{} & 49.5          & 48.6          & 20.7          & 20.6          \\
SpinNet\cite{ao2020spinnet}  & \multicolumn{1}{l}{} & 47.5          & 47.2          & 20.5          & 20.1          \\
Predator\cite{predator}      & \multicolumn{1}{l}{} & \underline{58.0}    & \underline{58.2}    & \textbf{26.7} & \underline{26.2}    \\
YOHO                         & \multicolumn{1}{l}{} & \textbf{64.4} & \textbf{65.1} & \underline{25.9}    & \textbf{26.4} \\ \hline

                             & \#Iters              & \multicolumn{4}{c}{Registration Recall (\%)}                  \\ \hline
RelativeNet\cite{deng20193d} & $\sim$1k             & 77.7          & -             & -             & -             \\
PerfectMatch\cite{smooth}    & 50k                  & 78.4          & 78.4          & 33.0          & 34.9          \\
FCGF\cite{FCGF}              & 50k                  & 85.1          & 84.8          & 40.1          & 40.8          \\
D3Feat\cite{d3feat}          & 50k                  & 81.6          & 83.0          & 37.2          & 36.1          \\
LMVD\cite{li2020end}         & 50k                  & 82.5          & 82.2          & 41.3          & 40.1          \\
EPN\cite{EPN}                & 50k                  & 88.2          & 87.6          & 58.1          & 58.9           \\
SpinNet\cite{ao2020spinnet}  & 50k                  & 88.6          & 88.4          & 59.8          & 58.1          \\
Predator\cite{predator}      & 50k                  & 89.0          & 88.4          & 59.8          & 57.7          \\
\multirow{2}{*}{YOHO-C}      & 0.1k                 & 90.1          & 89.4          & 62.9          & 64.2          \\
                             & 1k                   & \underline{90.5}    & \underline{90.6}    & \underline{64.9}    & 64.5          \\
\multirow{2}{*}{YOHO-O}      & 0.1k                 & 90.3          & 90.2          & 64.8          & \underline{65.5}    \\
                             & 1k                   & \textbf{90.8}          & \textbf{90.6} & \textbf{65.2} & \textbf{65.9} \\
\bottomrule[1.3pt]
\end{tabular}}
\end{center}
\caption{Results on the 3DMatch and 3DLoMatch datasets. The rotated version means that we adding additional arbitrary rotations to all point clouds. Same as \cite{ao2020spinnet,predator}, we set $\tau_{c}$=0.1m and $\tau_{r}$=0.2m to compute metrics.}
\label{tab:main}
\vspace{-20pt}
\end{table}

\begin{figure*}
\begin{center}
\includegraphics[width=1\linewidth]{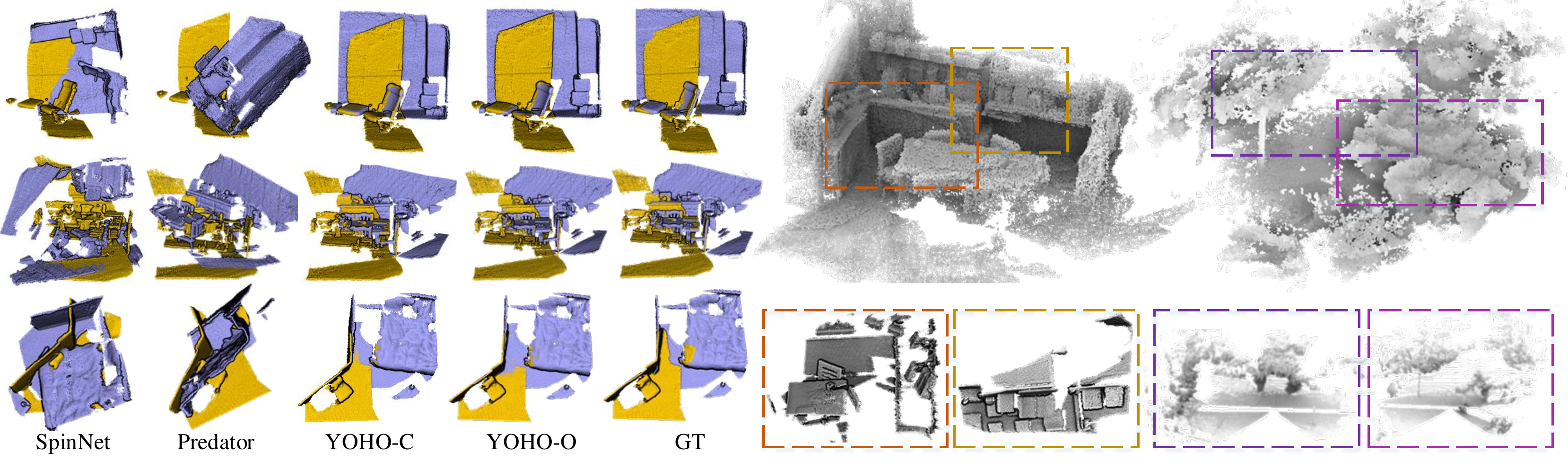}
\end{center}
\vspace{-10pt}
\caption{(Left) Qualitative comparison with baselines. (Right) Completed scenes by YOHO and some input partial scans.}
\label{fig:3dmatchresult}
\vspace{-5pt}
\end{figure*}

\subsection{Results on 3DMatch/3DLoMatch}
Results of YOHO and baseline models on the 3DMatch dataset and the 3DLoMatch dataset are shown in Table~\ref{tab:main}. Some qualitative results are shown in Fig.~\ref{fig:3dmatchresult}.
With 50 times fewer RANSAC iterations, YOHO still outperforms all baseline methods. 
The improvements on RR of YOHO mainly come from YOHO's utilization of rotation equivariance, which greatly reduces the searching space. 

\begin{table}[]
\begin{center}
\renewcommand{\arraystretch}{1.1}
\resizebox{\linewidth}{!}{
\begin{tabular}{lccccccc}
\toprule[1.5pt]
             &           & \multicolumn{3}{c}{$\tau_{c}=0.1m, \tau_{r}=0.2m$}       & \multicolumn{3}{c}{$\tau_c=0.2m, \tau_r=0.5m$} \\
             & \multicolumn{1}{c}{T(min)} & \multicolumn{1}{c}{FMR} & \multicolumn{1}{c}{IR} & \multicolumn{1}{c}{RR} & \multicolumn{1}{c}{FMR} & \multicolumn{1}{c}{IR} & \multicolumn{1}{c}{RR} \\ \hline
PerfectMatch\cite{smooth}   & 53.7        & \underline{79.2}     & \underline{13.3}    & 83.5          & \underline{96.8}    & 19.7          & 86.4                   \\
FCGF\cite{FCGF}             & 5.2         & 47.3           & 5.5           & 48.2          & 59.0          & 16.7          & 52.2                   \\
D3Feat\cite{d3feat}         & \textbf{4.1}& 59.1           & 7.9           & 55.2          & 63.3          & 18.2          & 59.1                   \\
SpinNet\cite{ao2020spinnet} & 62.6        & \textbf{92.0}  & \textbf{14.3} & \textbf{93.4} & \textbf{99.4} & \underline{23.2}    & \underline{96.0}             \\
Perdator\cite{predator}     & 16.7        & 25.4            & 3.7          & 52.3           & 65.6           & 11.1           & 74.7                    \\
YOHO-C                      & \underline{4.6}   & 71.1           & 10.6          & \underline{87.1}    & \underline{96.8}    & \textbf{26.8} & \textbf{96.8}          \\
YOHO-O                      & 6.2         & 71.1           & 10.6          & 74.1          & \underline{96.8}    & \textbf{26.8} & 94.7                   \\
\bottomrule[1.5pt]
\end{tabular}
}
\end{center}
\caption{Results on the ETH dataset. RANSAC is executed 1k iterations for YOHO and 50k for other methods. T is the total time for the registration on the ETH dataset.}
\label{tab:eth}
\vspace{-20pt}
\end{table}

\begin{table}[]
\begin{center}
\resizebox{\linewidth}{!}{
\begin{tabular}{lclllllll}
\toprule[1.5pt]
            &           & \multicolumn{7}{c}{$\tau_r(m)$}     \\ \cline{3-9} 
            & T(min)    & \multicolumn{1}{c}{0.05} & \multicolumn{1}{c}{0.1} & \multicolumn{1}{c}{0.15} & \multicolumn{1}{c}{0.2} & \multicolumn{1}{c}{0.25} & \multicolumn{1}{c}{0.3} & \multicolumn{1}{c}{0.5} \\ \hline
SpinNet\cite{ao2020spinnet}+ICP     &  66.8         & 92.1          & 96.4          & 96.2          & 96.7          & 96.7          & 96.8          & 96.8              \\
YOHO-C+ICP                          & \textbf{9.1}  & \textbf{92.3} & \textbf{97.0} & \textbf{97.5} & \textbf{97.5} & \textbf{97.5} & \textbf{97.6} & \textbf{97.8}  \\
YOHO-O+ICP                            &  11.2         & 89.5          & 94.1          & 94.7          & 94.9          & 95.1          & 95.4          & 95.9          \\ 
\bottomrule[1.5pt]
\end{tabular}
}
\end{center}
\caption{RR on the ETH dataset with ICP. T is the total time for the registration, including the time used in ICP.}
\label{tab:ethicp}
\vspace{-20pt}
\end{table}

\subsection{Results on the ETH dataset}
\label{sec:maineth}

In Table~\ref{tab:eth}, we provide results in two different thresholds for correspondences and registration and some qualitative results are shown in Fig.~\ref{fig:3dmatchresult}. In the strict setting with $\tau_c=0.1m$ and $\tau_r=0.2m$, YOHO underperforms SpinNet. The reason is that the FCGF backbone of YOHO downsamples input point clouds with a voxel size of 0.15m, which limits the accuracy of correspondences found by YOHO. SpinNet extracts local descriptors on every point independently, which has better accuracy but at a noticeable cost of longer computation time (62.6 min) than YOHO (4.6 min). In a slightly loose setting with $\tau_c=0.2m$ and $\tau_r=0.5m$, YOHO outperforms SpinNet in terms of IR and RR. 

Though the downsampled voxel size in FCGF makes YOHO unable to produce very accurate matches, we show that this can be easily improved by a commonly-used ICP~\cite{icp} post-processing. In Table~\ref{tab:ethicp}, we show RR using ICP post-processing, in which YOHO-C+ICP outperforms SpinNet+ICP in all thresholds including the strictest one with $\tau_r=0.05m$. Additional results on the WHU-TLS dataset can be found in the appendices.

\subsection{Ablation studies}
\label{sec:ablation}
To show the effectiveness of each component in YOHO, we conduct ablation studies on the 3DMatch dataset. The results are shown in Table~\ref{tab:ablation}. We design 5 models sequentially and every model only differs from the previous model on one component.

\textbf{Invariance via group features}. The model 0 in Table~\ref{tab:ablation} simply applies FCGF~\cite{FCGF} to construct descriptors. Based on the model 0, The model 1 achieves rotation invariance by constructing the icosahedral group features with the same FCGF and average-pooling on the group features. By comparing the model 1 with the model 0, we can see that achieving rotation invariance from group features is more robust. Further analysis on robustness against noise and density variations can be found in the appendices.

\textbf{Coarse rotation verification}. Based on the model 1, the model 2 estimates the coarse rotation from the group features $f_0$ and uses the coarse rotation verification (CRV) in the RANSAC. By comparing the model 2 with the model 1, CRV achieves higher RR than the vanilla RANSAC with 50 times fewer iterations.

\textbf{Group convolution layer}. Based on the model 2, the model 3 adds the proposed group convolutions before average-pooling, i.e. YOHO-C. The group convolution enables the network to exploit patterns defined on the icosahedral group, which brings about 4.9\% improvements on the FMR and 9.6\% improvements on the RR.

\textbf{One-shot transformation estimation}. Based on the model 3, the model 4 estimates the rotation residuals and uses the refined rotations to do the one-shot transformation estimation (OSE) in RANSAC, i.e. YOHO-O, which brings further improvements on RR.

\subsection{Analysis}
We provide more comparisons, analysis on iteration number and running time in the following. More analysis about robustness to noise, point density can be found in the appendices.

\begin{table}
\begin{center}
\resizebox{0.8\linewidth}{!}{
\begin{tabular}{ccccc|rr}
\toprule[1.3pt]
id & Inv. & GConv. & \#Iter & RANSAC & FMR & RR \\ 
\hline
 0 & None  &            & 50k & Vanilla  & 90.0 & 76.2 \\
 1 & Group &            & 50k & Vanilla  & 93.3 & 77.8 \\ 
 2 & Group &            & 1k & CRV       & 93.3 & 80.9 \\ 
 3 & Group & \checkmark & 1k & CRV       & 98.2 & 90.5 \\ 
 4 & Group & \checkmark & 1k & OSE       & 98.2 & 90.8 \\ 
\bottomrule[1.3pt]
\end{tabular}
}
\end{center}
\caption{Ablation studies on the 3DMatch dataset. The ``Inv." means how to get rotation invariance. ``None" means no rotation invariance while ``Group" means average-pooling on icosahedral group features. ``GConv" means the proposed group convolution. ``CRV" means coarse rotation verification while ``OSE" means one-shot rotation estimation. }
\label{tab:ablation}
\vspace{-25pt}
\end{table}

\textbf{Performance with different numbers of sampled points.}
Results in Table~\ref{tab:sample} show that YOHO consistently outperforms all other local descriptors in all cases but underperforms the learning-based matcher Predator~\cite{predator} when very few keypoints are used. The reason is that as a matcher, Predator is able to simultaneously utilize information from both source and target point clouds to find overlapped regions. However, YOHO only extracts descriptors on every point cloud separately, which is oblivious of the other point cloud to be aligned. Meanwhile, it is possible to incorporate YOHO within a learning-based matcher like Predator~\cite{predator} for better performance, which we leave for future works.

\begin{table}[]
\begin{center}
\renewcommand{\arraystretch}{1.3}
\resizebox{\linewidth}{!}{
\begin{tabular}{lcccccccccc}
\toprule[1.5pt]
          & \multicolumn{5}{c}{3DMatch}                                                                        & \multicolumn{5}{c}{3DLoMatch}                                                 \\
\#Samples & 5000          & 2500          & 1000          & 500           & 250                                & 5000          & 2500          & 1000          & 500           & 250           \\ \hline
\multicolumn{11}{c}{Feature Matching Recall (\%)}                                                                                                                                              \\ \hline
FCGF\cite{FCGF}      & 97.4          & \underline{97.3}          & \underline{97.0}          & \underline{96.7}          & \multicolumn{1}{c|}{\textbf{96.6}}          & 76.6          & 75.4          & 74.2          & 71.7          & 67.3          \\
D3feat\cite{d3feat}      & 95.6          & 95.4          & 94.5          & 94.1          & \multicolumn{1}{c|}{93.1}          & 67.3          & 66.7          & 67.0            & 66.7          & 66.5          \\
SpinNet\cite{ao2020spinnet}     & \underline{97.6}    &  97.2    & 96.8    & 95.5          & \multicolumn{1}{c|}{94.3}          & 75.3          & 74.9          & 72.5          & 70.0          & 63.6          \\
Predator\cite{predator}    & 96.6          & 96.6          & 96.5          & 96.3    & \multicolumn{1}{c|}{\underline{96.5}} & \underline{78.6}    & \underline{77.4}    & \textbf{76.3} & \textbf{75.7} & \textbf{75.3} \\
YOHO      & \textbf{98.2} & \textbf{97.6} & \textbf{97.5} & \textbf{97.7} & \multicolumn{1}{c|}{96.0}    & \textbf{79.4} & \textbf{78.1} & \textbf{76.3} & \underline{73.8}    & \underline{69.1}    \\ \hline
\multicolumn{11}{c}{Inlier Ratio (\%)}                                                                                                                                                        \\ \hline
FCGF\cite{FCGF}        & 56.8          & 54.1          & 48.7          & 42.5          & \multicolumn{1}{c|}{34.1}          & 21.4          & 20.0          & 17.2          & 14.8          & 11.6          \\
D3feat\cite{d3feat}      & 39.0          & 38.8          & 40.4          & 41.5          & \multicolumn{1}{c|}{\underline{41.8}}    & 13.2          & 13.1          & 14.0          & 14.6          & 15.0          \\
SpinNet\cite{ao2020spinnet}     & 47.5          & 44.7          & 39.4          & 33.9          & \multicolumn{1}{c|}{27.6}          & 20.5          & 19.0          & 16.3          & 13.8          & 11.1          \\
Predator\cite{predator}    & \underline{58.0}    & \underline{58.4}    & \textbf{57.1} & \textbf{54.1} & \multicolumn{1}{c|}{\textbf{49.3}} & \textbf{26.7} & \textbf{28.1} & \textbf{28.3} & \textbf{27.5} & \textbf{25.8} \\
YOHO      & \textbf{64.4} & \textbf{60.7} & \underline{55.7}    & \underline{46.4}    & \multicolumn{1}{c|}{41.2}          & \underline{25.9}    & \underline{23.3}    & \underline{22.6}    & \underline{18.2}    & \underline{15.0}    \\ \hline
\multicolumn{11}{c}{Registration Recall (\%)}                                                                                                                                                  \\ \hline
FCGF\cite{FCGF}        & 85.1          & 84.7          & 83.3          & 81.6          & \multicolumn{1}{c|}{71.4}          & 40.1          & 41.7          & 38.2          & 35.4          & 26.8          \\
D3feat\cite{d3feat}      & 81.6          & 84.5          & 83.4          & 82.4          & \multicolumn{1}{c|}{77.9}          & 37.2          & 42.7          & 46.9          & 43.8          & 39.1          \\
SpinNet\cite{ao2020spinnet}     & 88.6          & 86.6          & 85.5          & 83.5          & \multicolumn{1}{c|}{70.2}          & 59.8          & 54.9          & 48.3          & 39.8          & 26.8          \\
Predator\cite{predator}    & 89.0            & \underline{89.9}          & \textbf{90.6} & \underline{88.5} & \multicolumn{1}{c|}{\textbf{86.6}} & 59.8          & 61.2          & \underline{62.4}          & \textbf{60.8} & \textbf{58.1} \\
YOHO-C    & \underline{90.5}    & 89.7    & 88.4    & 87.6    & \multicolumn{1}{c|}{82.8}          & \underline{64.9}    & \underline{65.1}    & 61.4    & 54.5          & 43.9          \\
YOHO-O    & \textbf{90.8} & \textbf{90.3} & \underline{89.1}          & \textbf{88.6}          & \multicolumn{1}{c|}{\underline{84.5}}    & \textbf{65.2} & \textbf{65.5} & \textbf{63.2} & \underline{56.5}    & \underline{48.0}      \\ 
\bottomrule[1.5pt]
\end{tabular}}
\end{center}
\caption{Quantitative results on the 3DMatch and the 3DLoMatch datasets using different numbers of sampled points. RANSAC is executed 1k iterations for YOHO and 50k for other methods.}
\label{tab:sample}
\vspace{-20pt}
\end{table}

\begin{figure}
\begin{center}
\includegraphics[width=1\linewidth]{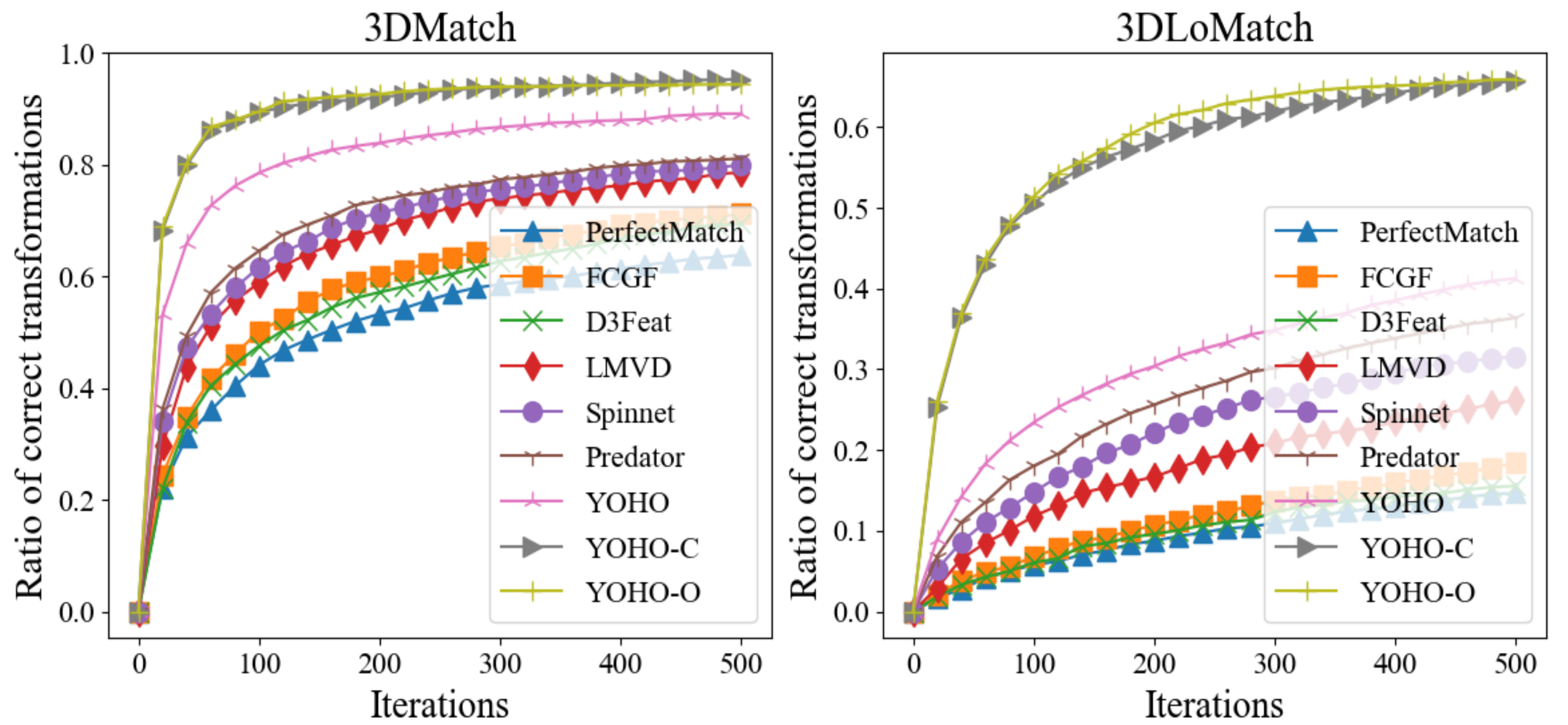}
\vspace{-15pt}
\end{center}
   \caption{Ratio of correct transformations versus iteration number on the 3DMatch dataset and the 3DLoMatch dataset.}
\label{fig:TR}
\vspace{-15pt}
\end{figure}

\textbf{Necessary iteration number}. To further show how many iterations are necessary to find a correct transformation, we further conduct an experiment on the 3DMatch/3DLoMatch datasets. For every scan pair, we count the number of iterations required to find a correct transformation. As shown in Fig.~\ref{fig:TR}, a point (R,N) in the figure means R\% scan pairs use less than N iterations to find the true transformation. 
The iteration ends once a true transformation is found while RANSAC chooses the transformation with a max inlier number. Thus, the curve only reveals correspondence quality but is not affected by the termination criteria of RANSAC. 

The results show that both YOHO-C and YOHO-O find true transformations very fast with less than 400 iterations. In comparison, all baseline methods only find a small portion of true transformations within 500 iterations, even though they can also achieve good RRs after 50k iterations as shown in Table~\ref{tab:main}. 
Moreover, an even more significant performance gap is shown on the 3DLoMatch dataset. The reason is that the smaller overlap brings a lower inlier ratio on the putative correspondences, which is less than 0.1 in general. In this case, selecting a triplet of inliers in baseline methods will have a probability less than $0.1^3=0.001$ while the probability of finding true transformations in YOHO-O is the same as the inlier ratio because it only needs one matched pair to compute a transformation.

\textbf{Running time}.
On a desktop with an i7-10700 CPU and a 2080Ti GPU, the time consumption is listed in Table~\ref{tab:time_new}. We provide the time $t_1$ used in the feature extraction of one point cloud fragment and the time $t_2$ used in aligning a point cloud pair by RANSAC, and the total time cost $T$ on registration of the 3DMatch and the 3DLoMatch dataset. The feature extraction in YOHO costs longer time than baselines because it needs to compute the backbone network 60 times. 
However, since these 60 point clouds are all rotated versions of the original one, they share many common computations like neighborhood querying for speeding up. 
Aligning scan pairs with RANSAC in YOHO is much faster than baselines. Meanwhile, the baselines uses advanced RANSAC implementation with engineering optimization, so YOHO with 1k iterations is not strictly 50 times faster than baselines with 50k iterations. In total, YOHO still takes the shortest time because there are 433 partial scans, 1623 scan pairs in 3DMatch and 1781 pairs in 3DLoMatch and we only need to extract features once and use them in all subsequent pair alignments.

\begin{table}[]\footnotesize
\renewcommand{\arraystretch}{1.02}
\begin{center}
\setlength{\tabcolsep}{3.0mm}{
\begin{tabular}{lcccc}
\toprule[1.0pt]
Method    & \multicolumn{1}{l}{\#Iters} & \multicolumn{1}{l}{$t_1$(s/pc)} & \multicolumn{1}{l}{$t_2$(s/pcp)}& \multicolumn{1}{l}{$T$(min)} \\
\hline
PerfectMatch\cite{smooth}    & 50k   & 22.125    & 0.368   & 180.547   \\
FCGF\cite{FCGF}              & 50k   & 0.381     & 0.384   & 24.535    \\
D3Feat\cite{d3feat}          & 50k   & 0.122     & 0.351   & 20.794    \\
EPN\cite{EPN}                & 50k   & 342.246    & 0.437   & 2494.668\\
SpinNet\cite{ao2020spinnet}  & 50k   & 26.556    & 0.413   & 215.077   \\
Predator\cite{predator}      & 50k   & -         & 1.221   & 69.271    \\
YOHO-C                       & 1k    & 1.812     & 0.056   & \textbf{16.253}    \\
YOHO-O                       & 1k    & 1.812     & 0.167   & 22.553    \\
\bottomrule[1.0pt]
\end{tabular}
}
\end{center}
\caption{The time consumption for the registration on the 3DMatch dataset and the 3DLoMatch dataset. We provide the time $t_1$ used in the feature extraction of one point cloud fragment and the time $t_2$ in aligning a point cloud pair, and the total time $T$ on the registration of the 3DMatch and the 3DLoMatch dataset.}
\label{tab:time_new}
\vspace{-25pt}
\end{table}

\begin{figure}
\begin{center}
\includegraphics[width=0.9\linewidth]{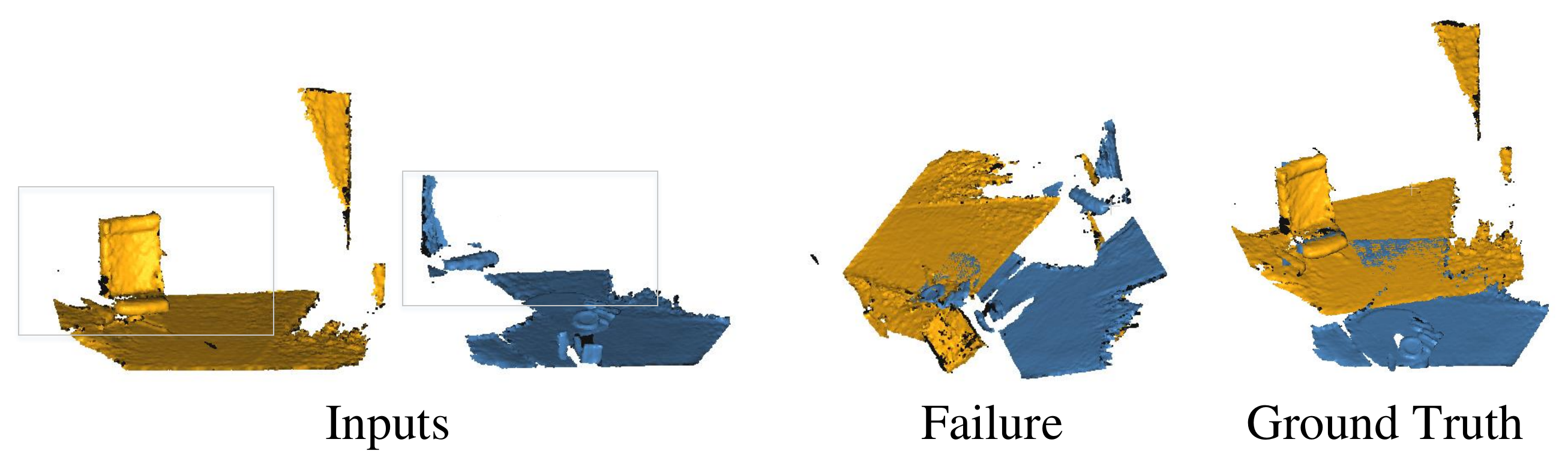}
\vspace{-10pt}
\end{center}
   \caption{A failure case on the 3DLoMatch dataset. The overlap area is planar, leading to ambiguity in rotation estimation.}
\label{fig:fail}
\vspace{-10pt}
\end{figure}

\textbf{Limitations}. 
The limitation of our methods mainly lies in two fronts: 1) When the overlap region of two scans mainly consists of planar points
, YOHO may fail to find the correct rotations since rotation estimation is ambiguous on planar points as in Fig.~\ref{fig:fail}.
2) Though YOHO is overall computation-efficient due to 
the improvement in RANSAC 
, the construction of YOHO-Desc is less efficient due to 60 times forward passes of backbone. 
This may possibly be optimized by group simplification~\cite{EMVN}, limiting rotations to SO(2), or advanced equivariance learning techniques~\cite{VN,tensorfield}, which we leave for future works.

\section{Conclusion}
In this paper, we propose a framework called YOHO for point cloud registration of two partial scans. The key of YOHO is a descriptor that simultaneously has rotation invariance and rotation equivariance. The descriptor is constructed by using the features defined on the icosahedral group which is rotation-equivariant by itself and can be pooled to achieve rotation invariance. We utilize the rotation-invariant part of YOHO-Desc to build correspondences and estimate a rotation per pair with their rotation-equivariant part. The estimated rotations greatly help the subsequent RANSAC to find the correct transformations robustly and accurately. We demonstrate the effectiveness of YOHO in multiple datasets, which achieves state-of-the-art performances with much fewer RANSAC iterations.
\section{Acknowledgement}
This research is jointly sponsored by the National Key Research and Development Program of China (No. 2018YFB2100503), the National Natural Science Foundation of China Projects (No. 42172431) and DiDi GAIA Research Collaboration Plan.

\clearpage
\bibliographystyle{ACM-Reference-Format}
\balance
\bibliography{sample-base}

\clearpage
\section{Appendices}
\appendix

\section{Proof of rotation equivariance}
\textbf{Rotation equivariance of $\bm{f}_0$}.
For an input neighboring point set $N_{\bm{p}}$, its group feature $\bm{f}_0$ can be written as:
\begin{equation}
    \bm{f}_0(g)={\varphi}(T_g\circ N_{\bm{p}})
    \label{eq:beforerotgf}
\end{equation}
where $\varphi$ is the FCGF backbone in YOHO.
Using a rotation $m \in G$ in the icosahedral group to rotate the input point set, the corresponding group feature will become:
\begin{equation}
    \bm{f}_0'(g)=\varphi(T_g\circ T_m\circ N_{\bm{p}})
    \label{eq:afterrotgf}
\end{equation}
Due to the closure property of a group, the composition $gm \in G$, Eq. \ref{eq:afterrotgf} can be expressed by
\begin{equation}
    \bm{f}_0'(g)=\varphi(T_{gm}\circ N_{\bm{p}})=\bm{f}_0(gm)
\end{equation}
Since $\bm{f}_0$ is represented by a $60\times n_0$ matrix and different row index represents different rotation $g\in G$, the $g$-th row on $\bm{f}'_0$ will be equal to the $gm$-th row of $\bm{f}_0$, which means $\bm{f}_0'$ is only a permuted version of $\bm{f}_0$, stated by
\begin{equation}
    \bm{f}_0'=P_m\circ \bm{f}_0
\end{equation}
where $P_m$ is a permutation operator of $m\in G$. 

\textbf{Rotation equivariance of $\bm{f}_k$}.
Here, we will prove that, if $\bm{f}_k$ is rotation-equivariant, which means $\bm{f}_{k}'=P_m\circ \bm{f}_k$, then the output feature $\bm{f}_{k+1}$ after the group convolution is also rotation-equivariant.
We have the following results.
\begin{equation}
    \begin{aligned}
    \left[\bm{f}_{k+1}'(g)\right]_j&=\sum_i^{13} \bm{w}_{j,i}^T \bm{f}_{k}'(h_i g)+b_j \\
                    &=\sum_i^{13} \bm{w}_{j,i}^T \bm{f}_{k}(h_i gm)+b_j \\
                    &=[\bm{f}_{k+1}(gm)]_j.
    \end{aligned}
\end{equation}
The first equation is the definition of the group convolution. The second equation holds because $\bm{f}'_{k}(g)=\bm{f}_k(gm)$. The third equation also uses the definition of the group convolution which treats $gm$ as a whole element. Since $\bm{f}_{k+1}'(g)=\bm{f}_{k+1}(gm)$ holds for any $g\in G$, $\bm{f}'_{k+1}=P_{m}\circ \bm{f}_{k+1}$.

\section{Implementation details}

\textbf{Training details for YOHO}. The backbone, Group Feature Embedder and Rotation Residual Regressor are trained sequentially. First, we train the backbone FCGF under the same setting as \cite{FCGF} for 80 epochs but only use the rotation data argumentation in $[0^{\circ},50^{\circ}]$. Then, we train the group feature embedder for 10 epochs with a batch size of 32 with all rotation data augmentation of all angles. The training uses the Adam optimizer with a learning rate of 1e-4. The learning rate is exponentially decayed by a factor of 0.5 every 1.8 epoch. Then, the following Rotation Residual Regressor is trained for 10 epochs with the same settings as the YOHO-Desc except that the origin learning rate is 1e-3 and the decay step is 3 epochs.

\textbf{Evaluation details}. 
All baselines are evaluated using their official codes and models. Only the estimated correspondences are used for the inlier counting in both two modified RANSAC, same as \cite{ao2020spinnet}.

\section{Comparison with direct registration methods}
We further compare YOHO with methods designed specifically for direct registration of partial point clouds, i.e.,  Go-ICP~\cite{yang2015go}, FGR~\cite{zhou2016fast},TEASER~\cite{yang2020teaser}, PointNetLK~\cite{aoki2019pointnetlk}, 3DRegNet~\cite{pais20203dregnet}, DGR~\cite{choy2020deep}, DHV~\cite{lee2021deep}, and PointDSC \cite{bai2021pointdsc}. Following ~\cite{lee2021deep,bai2021pointdsc}, we report the percentage of successful alignment on 3DMatch and 3DLoMatch datasets in Table.~\ref{tab:drm}, where one registration result is considered successful if the rotation error and translation error between the predicted and ground truth transformation are less than 15° and 0.3m. It can be observed that YOHO outperforms the direct registration methods on both datasets.
Note that YOHO, as a descriptor, is not in conflict with these methods. These direct registration methods aim to directly estimate transformations on vanilla correspondences while the idea of YOHO is to associate a rotation on every vanilla correspondence. It would be an interesting topic to design new direct registration methods to process such rotation-associated correspondences in future works. 

\begin{table}[h]\footnotesize
\begin{center}
\renewcommand{\arraystretch}{1.3}
\begin{tabular}{lc|c}
\toprule[1.0pt]
           & 3DMatch        & 3DLoMatch      \\ \hline
Go-ICP~\cite{yang2015go}     & 22.90          & -              \\
FGR~\cite{zhou2016fast}        & 78.56          & -              \\
TEASER~\cite{yang2020teaser}     & 85.77          & -              \\
PointNetLK~\cite{aoki2019pointnetlk} & 1.61           & -              \\
3DRegNet~\cite{pais20203dregnet}   & 77.76          & -              \\
DGR~\cite{choy2020deep}        & 86.50          & 50.20          \\
DHV~\cite{lee2021deep}        & 91.40          & 64.60
  \\
PointDSC~\cite{bai2021pointdsc}   & 93.28    & 61.50          \\
YOHO-C     & \underline{93.30}    & \underline{66.40} \\
YOHO-O     & \textbf{93.47} & \textbf{67.20} \\
\bottomrule[1.0pt]
\end{tabular}
\end{center}
\caption{Comparison with direct registration methods on the 3DMatch and the 3DLoMatch datasets. The
results in the table excluding ours are taken from PointDSC~\cite{bai2021pointdsc} and DHV~\cite{lee2021deep}.}
\label{tab:drm}
\vspace{-20pt}
\end{table}

\begin{table}[h]\tiny
\begin{center}
\resizebox{\linewidth}{!}{
\begin{tabular}{lcccc}
\toprule[0.6pt]
Dataset   & FMR (\%) & IR (\%) & \#Iters & RR (\%) \\ \hline
3DMatch   & 72.7     & 28.4    & 50k     & 48.3    \\
3DLoMatch & 28.2     & 5.7     & 50k     & 14.8    \\
ETH       & 53.8     & 11.4    & 50k     & 47.2    \\
\toprule[0.6pt]
\end{tabular}
}
\end{center}
\caption{Performance of the PointNet backbone. All the metrics are computed with $\tau_c$=0.1m and $\tau_r$=0.2m.}
\label{tab:pn}
\vspace{-20pt}
\end{table}

\begin{table}[h]
\begin{center}
\resizebox{\linewidth}{!}
{
\begin{tabular}{lccccccc}
\toprule[1.0pt]
\multirow{2}{*}{Dataset}   & FMR  & IR   & RR-C & RR-O & T-C                    & T-O                    \\
& (\%) & (\%) & (\%) & (\%) & (min) & (min) \\
\hline
3DMatch   & 94.7 & 37.5 & 83.1 & 83.4 & \multirow{2}{*}{142.8} & \multirow{2}{*}{151.4} \\
3DLoMatch & 59.2 & 11.3 & 48.8 & 50.1 &                        &                        \\
ETH       & 85.4 & 12.8 & 97.8 & 89.8 & 44.2                   & 46.6                   \\
\toprule[1.0pt]
\end{tabular}
}
\end{center}
\caption{Results of YOHO-PN with 1k RANSAC iterations. All the metrics are computed with $\tau_c$=0.1m and $\tau_r$=0.2m. `-C' means the results of YOHO-C, `-O' means the results of YOHO-O. T is the total time cost on the 3DMatch/3DLoMatch datasets or the ETH dataset.}
\label{tab:yoho_pn}
\vspace{-20pt}
\end{table}

\section{PointNet as Backbone}
\label{sec:pn}
We train YOHO with a backbone of a simple 10-layer PointNet (YOHO-PN) processing a randomly sampled 1024-point local patch with the same radius selection as~\cite{smooth,ao2020spinnet}. We follow the same training settings as above to train the model. To avoid the rotation ambiguities, we filter out the planar keypoints by a simple threshold ($\ge$0.03) on the minimum eigenvalue calculated by PCA~\cite{wold1987principal}.
The origin performances of the PointNet are shown in Table~\ref{tab:pn}. The results of YOHO-PN are shown in Table~\ref{tab:yoho_pn}.

From the results in Table~\ref{tab:pn} with Table~\ref{tab:yoho_pn} and the performance of baseline methods in the main paper, we can see that: (1) YOHO-PN brings great improvements to the simple PN backbone. (2) Even with a simple backbone, YOHO-PN already outperforms baselines \cite{smooth,FCGF,d3feat,li2020end} in terms of RR. (3) As discussed in the ETH result section, since YOHO-PN does not need to downsample the point cloud, YOHO-PN achieves better FMR/IR than YOHO-FCGF in the strict setting with $\tau_c=0.1m$ and $\tau_r=0.2m$, which is comparable to SpinNet. Meanwhile, in terms of RR, YOHO-PN-C outperforms all baselines including SpinNet in this strict setting. (4) As a local patch based descriptor, YOHO-PN still costs less time than other local patch based descriptors SpinNet~\cite{ao2020spinnet} and PerfectMatch~\cite{smooth}.

\begin{figure}[h]
\begin{center}
\includegraphics[width=1\linewidth]{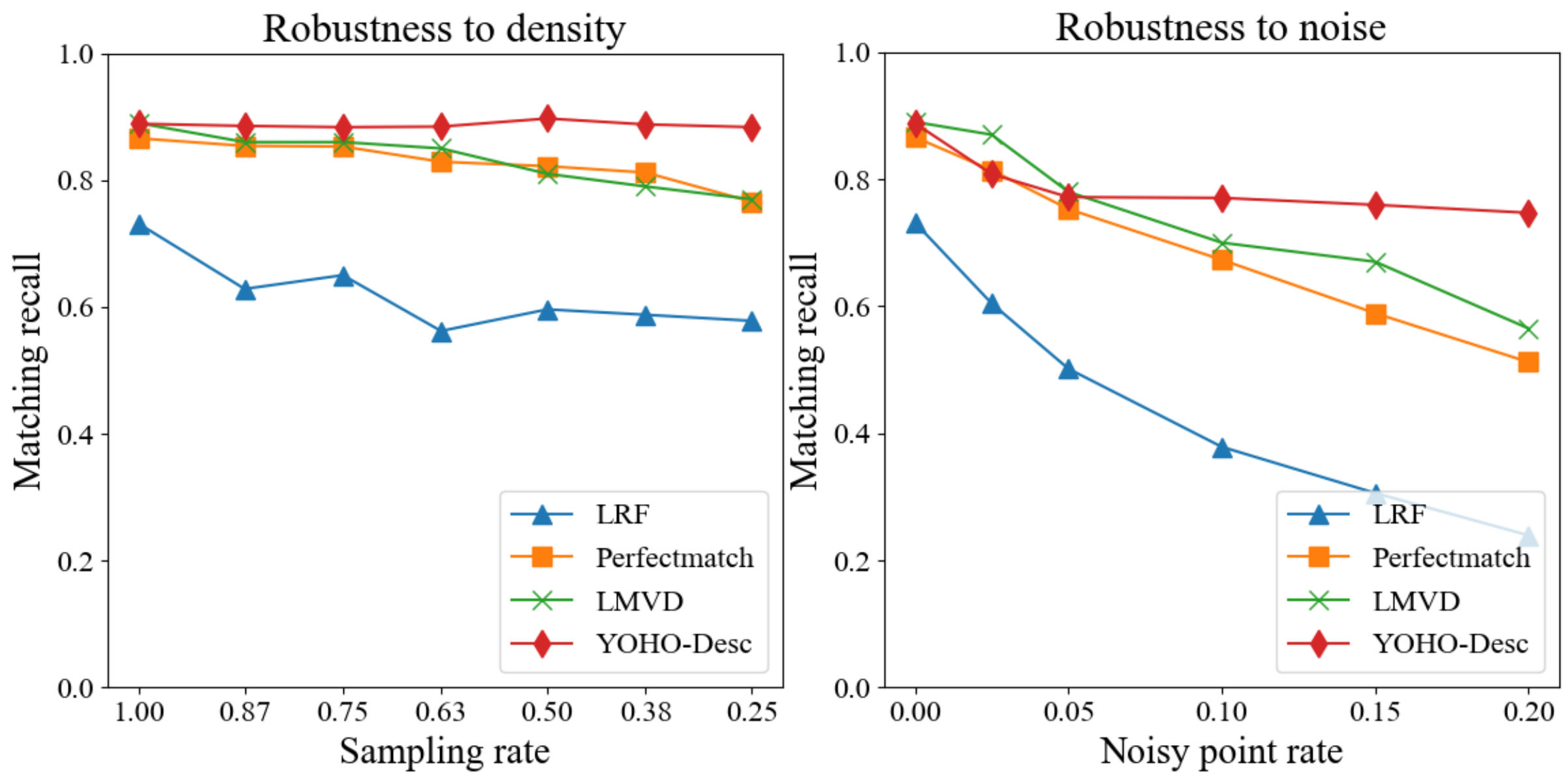}
\end{center}

   \caption{Ratio of correct patch pairs (y-axis) versus different levels of density and noise. In the experiment of density, we randomly drop some points in the patch and the x-axis shows the ratio of retained points. In the experiment of noise, we randomly add some noisy points to each patch where the x-axis shows the ratio of noisy points to the total point number.}
\label{fig:robust}
\vspace{-10pt}
\end{figure}

\section{Robustness analysis}
To further verify the robustness of YOHO to point density and noise, we conduct an experiment using YOHO-PN. We synthesize a dataset based on the 3DMatch dataset by randomly sampling 3000 patch pairs. Each of which contains 4096 points in a radius of 0.3m. We manually add random rotations and different levels of noise and density variations to every point cloud.

We compare the invariance from the proposed YOHO-PN with the (1) PCA+PointNet which achieves rotation invariance from PCA, (2) PerfectMatch~\cite{smooth} which achieves rotation invariance from plane fitting and (3) LMVD~\cite{li2020end} which achieves rotation invariance from multi-view images on the local point patch.
We report the ratio of correct patch pairs, i.e. matching recall, found by all descriptors.
The results are shown in Fig~\ref{fig:robust}. With the decreasing of the density and increasing of noise, the performance of YOHO-Desc almost stays the same or drops much slowly comparing with baselines, which demonstrates the robustness of our rotation-invariance construction.

\section{Results on the WHU-TLS dataset}
\label{sec:whu}
We further provide the results of our model on the WHU-TLS\cite{dong2020registration} dataset with $\tau_c=0.5m$ and $\tau_r=1m$ using 5000 keypoints, as shown in Table~\ref{tab:whu_whole}. 
For SpinNet\cite{ao2020spinnet}, we change the patch radius from 2m to 6m and choose the best results. For YOHO-PN, the patch radius is set to 6m. For YOHO, the downsample voxelsize is set to 0.8m. The results show that YOHO also generalizes better to the unseen WHU-TLS dataset than SpinNet.

\begin{table}[h]
\begin{center}
\resizebox{\linewidth}{!}{
\begin{tabular}{lccccccc}
\toprule[1.0pt]
Scenes    & Prk           & Mnt            & Cmp           & Riv           & Cav           & Tun           & Avg           \\ \hline
          & \multicolumn{7}{c}{Feature Matching Recall (\%)}                                                               \\ \hline
SpinNet\cite{ao2020spinnet}   & 22.6          & 0.0            & 0.0           & 0.0           & 36.4          & 0.0           & 9.8           \\
YOHO-PN   & \textbf{96.8} & \textbf{100.0} & \underline{33.3}    & \textbf{33.3} & \textbf{90.9} & \textbf{16.7} & \textbf{61.8} \\
YOHO      & \underline{93.5}    & \textbf{100.0} & \textbf{44.4} & \underline{16.7}    & \underline{81.8}    & \textbf{16.7} & \underline{58.9}    \\ \hline
          & \multicolumn{7}{c}{Inlier Ratio (\%)}                                                                         \\ \hline
SpinNet\cite{ao2020spinnet}   & 4.3           & 0.0            & 1.4           & 0.1           & 5.1           & 0.0           & 1.8           \\
YOHO-PN   & \textbf{13.6} & \underline{10.3}     & \underline{4.2}     & \textbf{5.0}  & \textbf{15.6} & \textbf{2.7}  & \textbf{8.6}  \\
YOHO      & \underline{13.2}    & \textbf{11.0}  & \textbf{5.5}  & \underline{4.0}     & \underline{11.2}    & \underline{1.4}     & \underline{7.7}     \\ \hline
          & \multicolumn{7}{c}{Registration Recall (\%)}                                                                   \\ \hline
SpinNet\cite{ao2020spinnet}       & 80.6          & 0.0            & 44.4          & 0.0           & 9.1           & 0.0           & 22.4          \\
YOHO-PN-C     & \textbf{90.3} & \textbf{100.0} & \textbf{88.9} & \textbf{16.7} & \underline{18.2}    & \underline{16.7}    & \textbf{55.1} \\
YOHO-PN-O     & 77.4          & 20.0           & 55.5          & 0.0           & 0.0           & \textbf{33.3} & 31.0          \\
YOHO-C        & \underline{87.1}    & \textbf{100.0} & \underline{77.8}    & 0.0           & \textbf{27.3} & \underline{16.7}    & \underline{51.5}    \\
YOHO-O        & 80.6          & \underline{80.0}     & 44.4          & 0.0           & 9.1           & \underline{16.7}    & 38.5          \\ \hline
YOHO-PN-C+ICP & 100           & 100            & 88.9          & 100           & 90.9          & 16.7          & \textbf{82.8} \\
YOHO-PN-O+ICP & 100           & 100            & 77.8          & 83.3          & 90.9          & 33.3          & 80.9          \\
YOHO-C+ICP    & 100           & 100            & 88.9          & 100           & 81.8          & 16.7          & \underline{81.2}    \\
YOHO-O+ICP    & 100           & 100            & 77.8          & 100           & 81.8          & 16.7          & 79.4          \\
\bottomrule[1.0pt]
\end{tabular}
}
\end{center}
\caption{Detailed results on the WHU-TLS dataset of different scenes.}
\vspace{-20pt}
\label{tab:whu_whole}
\end{table}

\section{Comparison with EPN~\cite{EPN}}
YOHO differs from EPN on two parts. The most different part is that when aligning partial scans like 3DMatch, YOHO propose a novel framework that takes use of the estimated rotation in the modified RANSAC to improve its accuracy and efficiency while EPN solely relies on rotation-invariant descriptors to build putative correspondences and adopts a vanilla RANSAC. The second difference is that EPN proposes a neat and efficient $SE(3)$ convolution layer, which decomposes $SE(3)$ into two groups and applies point convolutions and group convolutions in turns. 
However, EPN still relies features defined on the whole $SE(3)$ space, which means they need to retain the icosahedral group features on all points and every $SE(3)$ convolution layer involves icosahedral group features of all points. 
In comparison, YOHO first relies a backbone to extract the patterns defined on points and then we only consider the icosahedral group features for specific keypoints, which greatly reduce the computation burdens. 
Meanwhile, separating features extraction on points and icosahedral group also enables YOHO to adopt the computation-efficient backbone like FCGF~\cite{FCGF}. 
Our experiments demonstrate our superior performance over EPN especially in terms of computational efficiency.

\end{document}